\documentclass{article}

\usepackage{microtype}
\usepackage{graphicx}
\usepackage{subfigure}
\usepackage{booktabs} % for professional tables

\usepackage{hyperref}

\usepackage[accepted]{icml2024}

\usepackage{amsmath}
\usepackage{amssymb}
\usepackage{mathtools}
\usepackage{amsthm}

\usepackage[capitalize,noabbrev]{cleveref}

\theoremstyle{plain}

\theoremstyle{definition}

\theoremstyle{remark}

\usepackage[textsize=tiny]{todonotes}

\icmltitlerunning{Slot Abstractors}

\begin{document}

\twocolumn[
\icmltitle{Slot Abstractors: Toward Scalable Abstract Visual Reasoning}

% It is OKAY to include author information, even for blind
% submissions: the style file will automatically remove it for you
% unless you've provided the [accepted] option to the icml2024
% package.

% List of affiliations: The first argument should be a (short)
% identifier you will use later to specify author affiliations
% Academic affiliations should list Department, University, City, Region, Country
% Industry affiliations should list Company, City, Region, Country

% You can specify symbols, otherwise they are numbered in order.
% Ideally, you should not use this facility. Affiliations will be numbered
% in order of appearance and this is the preferred way.
% \icmlsetsymbol{equal}{*}

\begin{icmlauthorlist}
\icmlauthor{Shanka Subhra Mondal}{yyy}
\icmlauthor{Jonathan D. Cohen}{sch}
\icmlauthor{Taylor W. Webb}{comp}
% \icmlauthor{Firstname4 Lastname4}{sch}
% \icmlauthor{Firstname5 Lastname5}{yyy}
% \icmlauthor{Firstname6 Lastname6}{sch,yyy,comp}
% \icmlauthor{Firstname7 Lastname7}{comp}
%\icmlauthor{}{sch}
% \icmlauthor{Firstname8 Lastname8}{sch}
% \icmlauthor{Firstname8 Lastname8}{yyy,comp}
%\icmlauthor{}{sch}
%\icmlauthor{}{sch}
\end{icmlauthorlist}

\icmlaffiliation{yyy}{Department of Electrical and Computer Engineering, Princeton University, Princeton, NJ, US}
\icmlaffiliation{sch}{Princeton Neuroscience Institute, Princeton University, Princeton, NJ, US}
\icmlaffiliation{comp}{Department of Psychology, University of California Los Angeles, Los Angeles, CA, US}

\icmlcorrespondingauthor{Shanka Subhra Mondal}{smondal@princeton.edu}
% \icmlcorrespondingauthor{Firstname2 Lastname2}{first2.last2@www.uk}

% You may provide any keywords that you
% find helpful for describing your paper; these are used to populate
% the "keywords" metadata in the PDF but will not be shown in the document
\icmlkeywords{Machine Learning, ICML}

\vskip 0.3in
]

% this must go after the closing bracket ] following \twocolumn[ ...

% This command actually creates the footnote in the first column
% listing the affiliations and the copyright notice.
% The command takes one argument, which is text to display at the start of the footnote.
% The \icmlEqualContribution command is standard text for equal contribution.
% Remove it (just {}) if you do not need this facility.

\printAffiliationsAndNotice{}  % leave blank if no need to mention equal contribution
% \printAffiliationsAndNotice{\icmlEqualContributio} % otherwise use the standard text.

\begin{abstract}
Abstract visual reasoning is a characteristically human ability, allowing the identification of relational patterns that are abstracted away from object features, and the systematic generalization of those patterns to unseen problems. Recent work has demonstrated strong systematic generalization in visual reasoning tasks involving multi-object inputs, through the integration of slot-based methods used for extracting object-centric representations coupled with strong inductive biases for relational abstraction. However, this approach was limited to problems containing a single rule, and was not scalable to visual reasoning problems containing a large number of objects. Other recent work proposed Abstractors, an extension of Transformers that incorporates strong relational inductive biases, thereby inheriting the Transformer's scalability and multi-head architecture, but it has yet to be demonstrated how this approach might be applied to multi-object visual inputs. Here we combine the strengths of the above approaches and propose Slot Abstractors, an approach to abstract visual reasoning that can be scaled to problems involving a large number of objects and multiple relations among them. The approach displays state-of-the-art performance across four abstract visual reasoning tasks, as well as an abstract reasoning task involving real-world images.

\end{abstract}

\section{Introduction}

Abstract visual reasoning problems contain visual objects, the features of which collectively exemplify an abstract pattern or rule. These are a common testbed of human intelligence. Given a few example demonstrations of such problems, humans can easily identify the abstract pattern or rule, and then systematically generalize it to novel stimuli \cite{raven1938raven,kotovsky1996comparison,fleuret2011comparing,lake2015human}. Neural networks, on the other hand, often struggle to perform this type of abstract visual reasoning, tending to overfit to the concrete details of problems in the training data, and failing to extract the underlying abstract pattern or rule \cite{lake2018generalization,barrett2018measuring,ricci2018same}.

\begin{figure*}[h!]
\vskip 0.2in
\begin{center}
\centerline{\includegraphics[width=2.0\columnwidth]{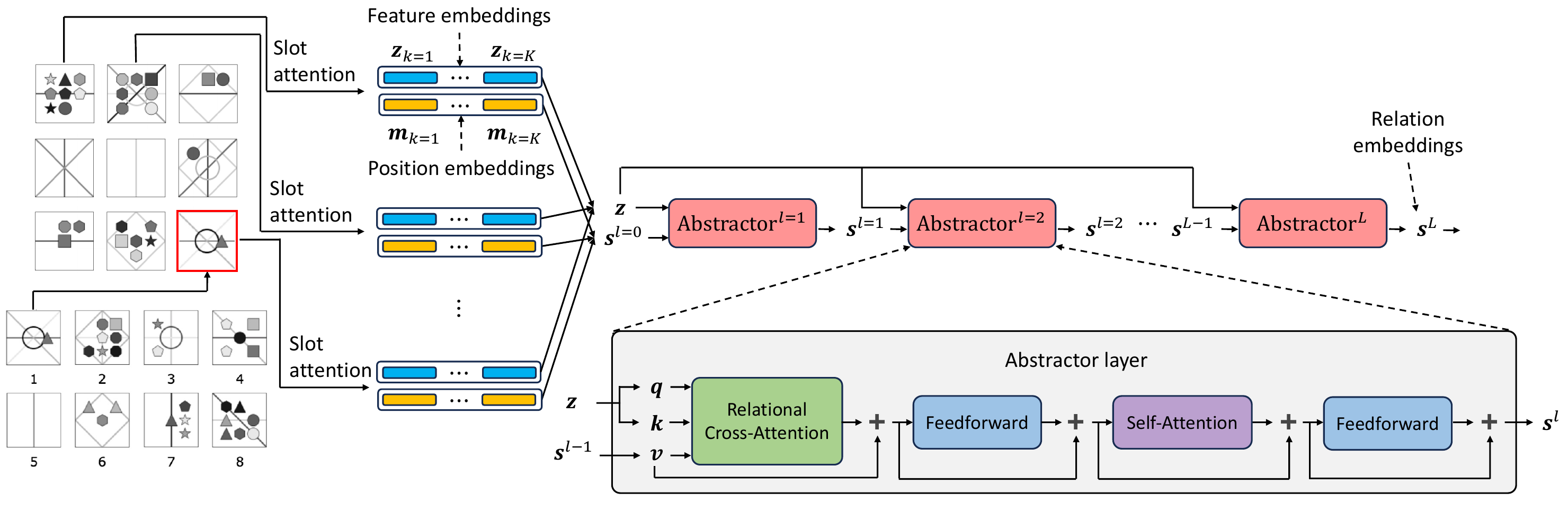}}
\caption{\textbf{Slot Abstractor}. The Slot Abstractor consists of two major components. First, object-centric representations are extracted using Slot Attention \cite{locatello2020object}. Relation embeddings are then computed using a series of Abstractor layers \cite{altabaa2023abstractors}. Example problem on left is from the PGM dataset \cite{barrett2018measuring}, consisting of a $3\times3$ matrix of image panels populated with objects. The task is to identify the abstract pattern among the image panels, and use this pattern to fill in the missing panel (bottom right), selecting from a set of eight choices. To generate scores for each answer choice, the corresponding image panel is inserted into the problem. Slot attention is then used to extract feature embeddings $\boldsymbol{z}_{k=1...K}$, and position embeddings $\boldsymbol{m}_{k=1...K}$ for each panel. Relation embeddings $\boldsymbol{s}$ are then computed through a series of Abstractor layers. Each layer consists of relational cross-attention, self-attention, and feedforward layers, with residual connections after each of these. Relational cross-attention uses feature embeddings to generate keys $k$ and queries $q$, and the relation embeddings from the previous layer to generate values $v$. Relation embeddings are initialized using position embeddings. After $L$ Abstractor layers, relation embeddings are averaged and passed through a linear layer to generate a score $y$.}
\label{OCARR_architecture}
\end{center}
\vskip -0.2in
\end{figure*}

To build models that can demonstrate human-like systematic generalization, several methods have recently been proposed \cite{webb2020learning,webb2020emergent,kerg2022neural,altabaa2023abstractors} to promote inference of relations among objects in the internal representations of neural networks. These methods implement a simple architectural inductive bias for relational abstraction, called the ``relational bottleneck" \cite{webb2023relational}, that drives the network to abstract over the features of objects, and identify relations among them that are necessary to perform the task. These methods demonstrated the learning and strong systematic generalization of abstract patterns from a few training examples, but were limited to inputs with pre-segmented visual objects. To address this, subsequent work proposed Object-Centric Relational Abstraction (OCRA) \cite{webb2023systematic}, an approach that integrates a relational bottleneck with object-centric representation learning methods \cite{greff2019multi,burgess2019monet,locatello2020object,engelcke2021genesis}. Though OCRA extends the capacity for strong abstract visual reasoning to images with more than one object, it is not scalable to problems with a large number of objects (example problem in \cref{OCARR_architecture}). This is because OCRA computes relational embeddings for all pairs of $N$ objects in a given scene, which are then further processed by a Transformer \cite{vaswani2017attention}, resulting in a combined complexity of $O(N^4)$. Moreover, OCRA's reasoning capacity was limited to problems involving a single relation.

Recently, Abstractors \cite{altabaa2023abstractors} were proposed as an extension of Transformers \cite{vaswani2017attention} for modeling relations between objects disentangled from the object features. The Abstractor implements a relational bottleneck using \textit{relational cross-attention}, a variant of the attention mechanisms in Transformers. Though this approach exhibits superior sample efficiency and improved generalization performance compared to other relational architectures \cite{shanahan2020explicitly,kerg2022neural}, it has yet to be applied to visual problems involving multi-object inputs \cite{fleuret2011comparing,barrett2018measuring,webb2023systematic}.  Here, we address this challenge, taking a step toward the development of scalable abstract visual reasoning algorithms that can be effectively applied to problems with a large number of objects and multiple relations among them. 

Specifically, we combine Slot Attention \cite{locatello2020object}, a method for unsupervised object-centric representation learning, with Abstractors \cite{altabaa2023abstractors}. The combined approach, Slot Abstractors (\cref{OCARR_architecture}), inherits the $O(N^2)$ complexity of Transformers, allowing it to be scaled to much more complex problems than previous abstract visual reasoning methods. The approach also inherits the Transformer's multi-head architecture, enabling it to solve problems involving multiple relations. We evaluate the Slot Abstractor on five abstract visual reasoning tasks (including a task involving real-world images), with a diverse range of visual and rule complexity. We find that the Slot Abstractor is capable of strong systematic generalization of learned abstract rules, and can be scaled to problems with multiple rule types and more than 100 objects, significantly improving over a number of competitive baselines in most settings.

\section{Approach}

\cref{OCARR_architecture} shows a schematic description of our approach (described in detail in \cref{alg:slot abstractors alg}). The Slot Abstractor consists of two major components: 1) extraction of object-centric representations; and 2) extraction of relational representations abstracted away from the object-level representations. We describe these two components in detail in the following sections.

\subsection{Object-Centric Representation Learning}

The Slot Abstractor employs slot attention \cite{locatello2020object} in the same way as OCRA \cite{webb2023systematic}, to extract object-centric representations from multi-object visual inputs. 

Given an image $\boldsymbol{x}$, slot attention learns to extract a set of latent embeddings (i.e., \textit{slots}), each of which captures a focused representation of a single visual object, when trained to reconstruct the image through decoding the object representations. The entire process is done without ground truth segmentation data and hence is completely unsupervised.

To accomplish this, the image is first passed through a convolutional encoder, which generates a feature map $\boldsymbol{feat}\in \mathbb{R}^{H \times W \times D}$. A position code (encoding the four cardinal directions) is passed through a linear layer, which generates positional embeddings $\boldsymbol{pos}\in \mathbb{R}^{H \times W \times D}$. Feature and position embeddings are additively combined, and then passed through a series of 1x1 convolutions, followed by flattening to generate $\boldsymbol{inputs}\in \mathbb{R}^{N \times D}$ (where $N = H \times W$). 

Then, a set of $K$ $\boldsymbol{slots}\in \mathbb{R}^{K \times D}$ is randomly initialized (from a normal distribution with learnable shared mean and variance parameters). The image is then presented, and the slots attend over the pixels of the feature map, through a transformer-style cross-attention. Specifically, each slot emits a query $\operatorname{q}(\boldsymbol{slots}) \in \mathbb{R}^{K\times D}$ (through a linear projection), and each location in the feature map emits a key $\operatorname{k}(\boldsymbol{inputs}) \in \mathbb{R}^{N \times D}$ and a value $\operatorname{v}(\boldsymbol{inputs}) \in \mathbb{R}^{N \times D}$. An attention distribution over the feature map $\boldsymbol{attn} = \operatorname{softmax} (\frac{1}{\sqrt{D}} \operatorname{k}(\boldsymbol{inputs}) \cdot \operatorname{q}(\boldsymbol{slots})^{\top})$ is generated for each slot, and a weighted mean of the values $\boldsymbol{updates} = \boldsymbol{attn} \cdot \operatorname{v}(\boldsymbol{inputs})$ is used to update the slot representations using a Gated Recurrent Unit \cite{cho2014learning}, followed by a feedforward network with residual connection. After $T$ iterations of slot attention, the updated slot representations
are passed through a spatial broadcast decoder  \cite{watters2019spatial}, which generates both a reconstructed image and a mask for each slot. The masks are normalized using a softmax across the slots, and the normalized masks are used to compute a weighted average of the slot-specific reconstructions, which generates a combined reconstruction of the input image.
 
To compute relational representations, it is crucial to represent position distinctly from other object features. For a given input image, we used the final attention map $\boldsymbol{attn}_{T} \in \mathbb{R}^{K\times N}$ after $T$ iterations of slot attention, to compute feature- and position-specific embeddings for each slot:

\begin{equation}
\label{z_k_eq}
    \boldsymbol{z}_{k} = \boldsymbol{attn}_{T} \operatorname{flatten}(\boldsymbol{feat})[k]
\end{equation}

\begin{equation}
\label{m_k_eq}
    \boldsymbol{m}_{k} = \boldsymbol{attn}_{T} \operatorname{flatten}(\boldsymbol{pos})[k]
\end{equation}

\noindent where $\boldsymbol{feat}$ and $\boldsymbol{pos}$ represent the feature- and position-embeddings before being additively combined. The position embeddings help to keep track of which relations correspond to which objects.

Although slot attention works well for synthetic images, it is not as effective for real-world images \cite{seitzer2022bridging}. To extend our approach to problems involving real-world images, we used DINOSAUR \cite{seitzer2022bridging}, an approach in which slot attention is applied to the features obtained from DINO \cite{caron2021emerging} (a large-scale pre-trained vision transformer (ViT) \cite{dosovitskiy2020image}). This approach has been shown to be an effective method for unsupervised instance segmentation of real-world images. The final attention map is used with the ViT's positional embeddings to compute position-specific embeddings for each slot as described in \cref{m_k_eq}. The slot embeddings obtained after performing slot attention over the ViT's features constituted the feature-specific embeddings for each slot.

 We assume that human visual representations are generally not learned from scratch when performing each new task. Instead, visual object representations are shaped by a wide variety of non-task-specific prior experiences. To model the effects of this prior experience, we pre-trained slot attention on a reconstruction objective, using a slot-based autoencoder framework (more details in \cref{expt_details}).

\subsection{Relational Representation Learning}
\label{abstractor}
The Slot Abstractor employs the Abstractor module \cite{altabaa2023abstractors} to process the object-centric representations extracted by the Slot Attention mechanism. The core operation of the Abstractor module is relational cross-attention ($\operatorname{RCA}$), a variant on the standard form of cross-attention found in Transformers \cite{vaswani2017attention}. Like Transformers, the Abstractor also employs a multi-head architecture, enabling it to model problems with multiple distinct relations. 

The Abstractor module computes abstract relational representations $\boldsymbol{s}$ over a series of $L$ layers. The initial relational representations $\boldsymbol{s}^{l=0}$ are formed using the position-specific embeddings $\boldsymbol{m}$ obtained from slot attention. Then, at each layer $l$, these representations are updated using multi-head relational cross-attention between the feature-specific embeddings $\boldsymbol{z}$:

\begin{equation}
\boldsymbol{s}^{l} = \operatorname{RCA}(\boldsymbol{z}, \boldsymbol{s}^{l-1})
\end{equation}

\begin{equation}
\operatorname{RCA}(\boldsymbol{z}, \boldsymbol{s}^{l-1}) = \operatorname{concat}(\boldsymbol{\tilde{s}^{h=1}},..,\boldsymbol{\tilde{s}^{h=H}})\boldsymbol{W_o}
\end{equation}

\begin{equation}
\boldsymbol{\tilde{s}}^{h} = \operatorname{softmax}\bigg(\frac{(\boldsymbol{z}\boldsymbol{W_q}^{h})^T (\boldsymbol{z}\boldsymbol{W_k}^{h})}{\sqrt{D}} \bigg) \boldsymbol{s}^{l-1}\boldsymbol{W_v}^{h}
\label{symbol_eq}
\end{equation}
    
where $\boldsymbol{W_q}^{h} \in \mathbb{R}^{D \times D}, \boldsymbol{W_k}^{h} \in \mathbb{R}^{D \times D}, \boldsymbol{W_v}^{h} \in \mathbb{R}^{D \times D}$ are the linear projection matrices used by the $h^{\text{th}}$ head to generate queries, keys, and values respectively; $\boldsymbol{\tilde{s}}^{h}$ is the result of relational cross-attention in the $h^{\text{th}}$ head; and $\boldsymbol{W_o}$ are the output weights through which the concatenated outputs of all $H$ heads are passed. The key difference relative to standard cross-attention is that queries and keys are generated from the input features, enabling the relations between those features to be modeled through inner products, and the values are generated from a separate set of (position-specific) embeddings. This relational variant on cross-attention implements a relational bottleneck \cite{webb2023relational}, in that downstream processing (which only has access to $\boldsymbol{s}$) is driven only by the relations between objects, as represented by inner products (see Altabaa and Lafferty \yrcite{altabaa2024approximation} for theoretical considerations regarding the use of inner products to represent relations), and thus is disentangled from the individual features of those objects. We hypothesized that this relational bottleneck would enable systematic generalization to problems with previously unseen object features.

In each layer, multi-head relational cross-attention was composed together with feedforward networks and standard multi-head self-attention, with residual connections after each of these, as depicted in Figure~\ref{OCARR_architecture}, and described in detail in \cref{alg:slot abstractors alg}.  The application of multiple layers of relational and self-attention enabled the Slot Abstractor to flexibly model higher-order relations (relations between relations), which play an important role in abstract visual reasoning. Crucially, this approach retains the quadratic complexity of standard Transformers (given $K$ slots, Slot Abstractors have $O(K^2)$ complexity), making it feasible to scale the approach to visual inputs containing a large number of objects.
After $L$ layers, the relational representations were averaged over the $K$ slots, and passed through a task-specific output layer to generate a score $y$. Depending on the task, a score $y$ was generated for each answer choice, and the abstractor module was trained using cross entropy loss. More details can be found in the last paragraph of \cref{expt_details}.

\section{Related Work}

A number of neural network models have been proposed \cite{barrett2018measuring,steenbrugge2018improving,zhang2019learning,zheng2019abstract,spratley2020closer,jahrens2020solving,wang2020abstract,wu2020scattering,benny2021scale,hu2021stratified,zhuo2022effective,zhang2022learning,yang2023neural} to solve abstract visual reasoning problems \cite{barrett2018measuring,zhang2019raven,teney2020v} based on Raven's Progressive Matrix (RPM) \cite{raven1938raven}. Many of these models incorporate problem-specific inductive biases, and therefore cannot be applied to other types of visual reasoning problems. To address this, the Slot Transformer Scoring Network (STSN) was proposed~\cite{mondal2023stsn}, combining a transformer reasoning module with a general-purpose inductive bias for object-centric visual processing~\cite{greff2019multi,burgess2019monet,locatello2020object,engelcke2021genesis,dittadi2021generalization,jiang2023object}. This approach achieved state-of-the-art performance on RPM-based abstract visual reasoning tasks, consistent with results from object-centric models in other visual reasoning tasks~\cite{ding2021attention,wu2022slotformer}. However, unlike human abstract reasoning, these approaches generally require very large training sets, and tend to generalize poorly out-of-distribution.

To address these limitations, neural network architectures have recently been proposed that incorporate strong relational inductive biases~\cite{webb2020emergent,kerg2022neural}. By constraining downstream processing to focus only on the relations between inputs (and to ignore the concrete features of those inputs), these architectures demonstrated human-like systematic generalization of learned abstract rules from a few training examples. This approach has more generally been referred to as the ``relational bottleneck" principle \cite{webb2023relational}. Some of these architectures, however, have been limited by their dependence on inputs consisting of pre-segmented visual objects. 

One recent approach, Object-Centric Relational Abstraction (OCRA)~\cite{webb2023systematic}, combined both strong relational and object-centric inductive biases (specifically Slot Attention~\cite{locatello2020object}) to demonstrate systematic generalization for abstract visual reasoning problems based on multi-object inputs. This approach was limited, however, both by its inability to process problems with multiple relations, and by its computational complexity ($O(N^4)$), preventing it from being scaled to problems containing a large number of objects. 

 Another recent work proposed Abstractors \cite{altabaa2023abstractors}, an extension of Transformers \cite{vaswani2017attention} that implemented the relational bottleneck using a relational variant of cross-attention. This approach demonstrated strong sample efficiency and generalization performance, but has yet to be explored for abstract visual reasoning problems with multi-object inputs \cite{fleuret2011comparing,barrett2018measuring,webb2023systematic}. Abstractors inherit the quadratic complexity of Transformers, making it feasible to scale them to problems with a large number of objects. Abstractors also employ a multi-head architecture, similar to Transformers, allowing them to model multiple distinct relations. Here we combine an unsupervised object-centric encoding mechanism, Slot Attention \cite{locatello2020object} with Abstractors, enabling abstract visual reasoning to be scaled to problems involving a large number of objects and multiple relations.

Though our proposed model employed previously designed components (slot attention and relational cross-attention), combining these components into an effective architecture necessitated many novel design choices. We have proposed a factorized slot representation, in which the position embeddings are used for the initial relational representations $\boldsymbol{s}^{l=0}$, whereas the feature embeddings are used to compute the keys and queries for the relational cross attention. We also find that relational cross attention was most effective when it was interleaved with self attention and feedforward layers in a particular manner. In \cref{ablation_study}, we show through ablation experiments targeted toward the use of the factorized slot representation (Row 4 of \cref{art-ablation-table}), and the interleaved relational-cross-attention/self-attention (Row 3 of \cref{art-ablation-table}), that these design choices are essential for obtaining strong performance. Thus, the novel contribution of our work is to identify an effective way for combining object-centric and relational components, yielding an architecture that achieves state-of-the-art performance on challenging abstract reasoning tasks.
\section{Experiments}

\subsection{Datasets}

We evaluated the Slot Abstractor on five challenging abstract visual reasoning datasets, ART \cite{webb2020emergent}, SVRT \cite{fleuret2011comparing}, CLEVR-ART \cite{webb2023systematic}, PGM \cite{barrett2018measuring}, and V-PROM \cite{teney2020v}. Problems in ART and SVRT datasets consist of simple 2D shapes, whereas CLEVR-ART consists of more complicated and realistic 3D shapes. Each problem of ART, SVRT, and CLEVR-ART contains a small number of objects and is governed by a single rule among them. Problems from the PGM dataset contain a much larger number of objects, with multiple rules among them. V-PROM is a matrix reasoning task (similar to PGM) involving real-world images. The datasets are described in more detail below. 

\subsubsection{ART}

The Abstract Reasoning Tasks (ART) dataset was proposed by Webb et al. \yrcite{webb2020emergent}. It consists of four visual reasoning tasks (`same/different', `relational-match-to-sample', `distribution-of-3', `identity rules'), each defined by a different abstract rule (\cref{art_dataset}). This dataset was created using 100 unicode character objects, and generalization regimes of varying difficulty were created defined by the number of unique objects used to instantiate the rules during training. We focused on the most difficult generalization regime, in which the training set consists of problems that are created using 5 out of the 100 possible objects, and the test problems are created using the remaining 95 objects. This is a difficult test of systematic generalization, as it requires learning of an abstract rule from a very small set of examples (\cref{art_examples}), with little perceptual overlap between training and test sets. Consistent with previous work \cite{webb2023systematic} we investigated a version of the tasks with multi-object displays and random spatial jitter (random translation of up to 5 pixels in any direction) applied to each object (which has previously been shown to increase task difficulty for some relational architectures~\cite{vaishnav2023gamr}).

\subsubsection{SVRT}

The Synthetic Visual Reasoning Test (SVRT) dataset \cite{fleuret2011comparing} consists of 23 binary classification tasks. Each task consists of synthetic 2D shapes, with an underlying abstract pattern among the shapes, and can be broadly grouped into two categories: those that are defined by same/different relations (\cref{svrt_sd_example}), and those that are defined by spatial relations (\cref{svrt_sr_example}). We used a small training set of 500 or 1000 examples per task, consistent with previous work~\cite{vaishnav2023gamr}. 

\subsubsection{CLEVR-ART}

This dataset (\cref{clevr_art_dataset}) was proposed by Webb et al. \yrcite{webb2023systematic}, and created using photorealistic synthetic 3D shapes from CLEVR \cite{johnson2017clevr}. It consists of two visual reasoning tasks from ART:  relational-match-to-sample and identity rules. The training set consists of images created using small and medium-sized rubber cubes in four colors (cyan, brown, green, and gray). The test set consists of images created using large-sized metal spheres and cylinders in four other colors (yellow, purple, blue, and red). The features of objects in the training and test set are completely different, which tests the systematic generalization of learned abstract rules to previously unseen object features.

\subsubsection{PGM}

The Procedurally Generated Matrices (PGM) dataset was proposed by Barrett et al. \yrcite{barrett2018measuring}, and is also based on Raven's Progressive Matrices problem sets \cite{raven1938raven}. Each problem consists of a $3\times3$ matrix of image panels populated with objects of varying shape, size, and color (\cref{pgm_dataset}). The task is to identify the abstract pattern among the image panels in the first two rows and/or columns of the matrix, and use that pattern to fill the panel in the third row and column from a set of eight choices. The maximum number of objects possible in an image panel is 16, with 9 panels per problem, yielding a total maximum of 144 objects per problem (compared to a maximum of 6 objects per problem in the other three datasets). Each matrix problem in PGM is defined by the abstract structure $\mathcal{S} = \{[r, o ,a]: r \in \mathcal{R}, o \in \mathcal{O}, a \in \mathcal{A} \}$, where $\mathcal{R} = $ \{progression, XOR, AND, OR, consistent union\} are the set of rules, $\mathcal{O}=$ \{shape, line\} are the set of objects, and $\mathcal{A}=$ \{size, type, position, color, number\} are the set of attributes. Problems consist of multiple rules (up to 4) governed by triples $[r, o ,a]$.

The PGM specifies eight different generalization regimes of varying difficulty. Each regime consists of 1.2M training problems, 20K validation problems, and 200K testing problems. The neutral regime is the easiest generalization regime, with training and test sets sampled from the same distribution. The other seven regimes --- interpolation (Intp.), extrapolation (Extp.), heldout attribute shape color (H.O.S-C), heldout attribute line type (H.O.L-T), heldout triples (H.O.Triples), heldout triple pairs (H.O.T.P.), heldout arribute pairs (H.O.A.P.) --- test out-of-distribution (OOD) generalization (see \cref{pgm_ood_regimes} for more details).

\subsubsection{V-PROM}

The Visual Progressive Matrices (V-PROM) dataset was proposed by Teney et al. \yrcite{teney2020v}. V-PROM is a matrix reasoning dataset (similar to PGM), but unlike standard matrix reasoning tasks, the problems in V-PROM are constructed from real-world images (\cref{vprom_dataset}). The maximum number of objects possible in an image is 10, and since there are 9 panels per problem, the total maximum number of objects is 90. Like PGM, each problem consists of multiple abstract rules and the dataset consists of generalization regimes of varying difficulty. In this work, we focused only on the Neutral regime, with around 139K training problems, 8K validation problems, and 73K test problems.

\begin{table*}[t]
\caption{Results on the four tasks of the ART dataset. Results reflect test accuracy averaged over 10 trained networks($\pm$ standard error). }
\label{art-table}
\vskip 0.15in
\begin{center}
\begin{small}
\begin{sc}
\begin{tabular}{lccccr}
\toprule
Model & Same/Different & Relational-match-to-sample  & Distribution-of-3 & Identity rules \\
\midrule
ResNet &66.60$\pm$1.5&49.89$\pm$0.2&50.07$\pm$1.3&54.84$\pm$2.4\\
Slot-CorelNet &50.50$\pm$0.2&49.82$\pm$0.2&26.80$\pm$0.8&43.50$\pm$5.2\\
Slot-ESBN &50.02$\pm$0.2&49.99$\pm$0.2&25.56$\pm$0.1&50.33$\pm$2.8\\
Slot-GAMR &62.98$\pm$1.4&59.55$\pm$2.7&32.77$\pm$1.0&61.92$\pm$0.9\\
Slot-RN &77.26$\pm$1.9&61.62$\pm$1.1&52.10$\pm$0.7&65.96$\pm$1.1\\
Slot-IN &59.23$\pm$2.3&56.93$\pm$0.8&49.48$\pm$1.8&72.82$\pm$1.6\\
Slot-Transformer   & 68.46$\pm$ 2.0& 73.99$\pm$ 3.0 & 60.61$\pm$ 1.9 & 78.32$\pm$ 1.8 \\
GAMR   & 83.49$\pm$ 1.4& 72.20$\pm$ 3.0 & 68.62$\pm$ 1.8 & 66.23$\pm$ 4.8 \\
OCRA   & 87.95$\pm$ 1.3& 85.31$\pm$ 2.0 & 86.42$\pm$ 1.3 & 92.8$\pm$ 0.3 \\
Slot-Abstractor &\textbf{96.36$\pm$ 0.4}& \textbf{91.64$\pm$ 1.6}&\textbf{95.22$\pm$ 0.4}&\textbf{96.41$\pm$0.4} \\
\bottomrule
\end{tabular}
\end{sc}
\end{small}
\end{center}
\vskip -0.1in
\end{table*}

\subsection{Baselines}

For the ART, SVRT, and CLEVR-ART datasets, we compared the Slot Abstractor to baseline methods reported in Webb et al. \yrcite{webb2023systematic}, which includes the OCRA model, the GAMR architecture proposed by Vaishnav and Serre \yrcite{vaishnav2023gamr}, ResNet-50 \cite{he2016deep}, a version of ResNet that uses self-attention (Attn-ResNet) \cite{vaishnav2022understanding}, and a set of baselines that combined pre-trained slot attention with various reasoning architectures (GAMR, Transformer, ESBN \cite{webb2020emergent}, CorelNet \cite{kerg2022neural}, Relation Net (RN) \cite{santoro2017simple}, and Interaction Net (IN) \cite{watters2017visual}).

For the PGM dataset, we compared the Slot Abstractor to the PredRNet model \cite{yang2023neural} and several state-of-the-art methods reported in Zhang et al. \yrcite{zhang2022learning}, including their proposed ARII model. Since our main focus was on systematic generalization, for comparison we only included models that were evaluated on at least one OOD regime. For the V-PROM dataset, we compared the Slot Abstractor to the best performing model (RN) reported in Teney et al. \yrcite{teney2020v}. We didn't use any auxiliary information (i.e., rule labels), and hence for a fair comparison we only compared to baselines that were trained without auxiliary loss. 

\begin{table*}[t]
\caption{Results on the two task categories of the SVRT dataset. Results reflect test accuracy averaged over different tasks from each category ($\pm$ standard error), for 1 trained network for each task. }
\label{svrt-table}
\vskip 0.15in
\begin{center}
\begin{small}
\begin{sc}
\begin{tabular}{lcccccr}
\toprule
Model & \multicolumn{2}{c}{Same/Different} &
  \multicolumn{2}{c}{Spatial Relations} \\
& Dataset Size =0.5k & Dataset Size =1k &  Dataset Size =0.5k & Dataset Size = 1k & \\
\midrule
ResNet
&54.97$\pm$2.2&56.88$\pm$2.5&85.18$\pm$4.3&94.87$\pm$1.6\\
Attn-ResNet &62.30$\pm$3.5&68.83$\pm$4.4&94.80$\pm$1.4&97.66$\pm$0.7\\
Slot-CorelNet &52.95$\pm$1.4&57.13$\pm$2.6&60.95$\pm$3.7&74.59$\pm$3.7 \\
Slot-ESBN &53.83$\pm$1.1&51.67$\pm$1.1&61.30$\pm$2.3&62.69$\pm$2.4\\
Slot-GAMR &63.06$\pm$3.7&66.87$\pm$3.2 &84.90$\pm$2.4&86.99$\pm$2.2\\
Slot-RN &71.48$\pm$4.8&81.79$\pm$4.4&91.73$\pm$1.8&96.20$\pm$1.4\\
Slot-IN &68.23$\pm$4.8&74.99$\pm$4.9&90.23$\pm$2.0&94.86$\pm$1.4\\
Slot-Transformer   & \textbf{76.54$\pm$5.1}& \textbf{89.85$\pm$4.2} & 94.06$\pm1.6$ & \textbf{97.86$\pm$0.9} \\
GAMR   & \textbf{76.80$\pm$4.9} & 82.05$\pm$4.4  & \textbf{97.40$\pm$0.7} & \textbf{98.74$\pm$0.3} \\
OCRA   & \textbf{79.89$\pm$4.5}& \textbf{90.30$\pm$4.1} & 89.25$\pm$2.5 & 95.02$\pm$2.4 \\
Slot-Abstractor &\textbf{82.20$\pm$4.7}&\textbf{91.86$\pm$4.0}&91.74$\pm$2.2&97.26$\pm$1.1\\
\bottomrule
\end{tabular}
\end{sc}
\end{small}
\end{center}
\vskip -0.1in
\end{table*}

\subsection{Experimental Details}
\label{expt_details}

We pre-trained slot attention using a reconstruction (i.e., autoencoding) objective. For ART, SVRT, and CLEVR-ART datasets we used the pre-trained model from Webb et al. \yrcite{webb2023systematic}\footnote{\href{https://github.com/Shanka123/OCRA}{https://github.com/Shanka123/OCRA}}. For PGM and V-PROM, we pre-trained slot attention on the neutral regime training set. \cref{cnn_enc_pgm} and \cref{slot_dec_pgm} describe the hyperparameters for the convolutional encoder and the slot-based spatial broadcast decoder respectively, whereas the pre-training hyperparameters are described in \cref{slot_attention_pretraining_hyperparams} for the PGM dataset. For the V-PROM dataset, \cref{cnn_enc_vprom} describes the hyperparameters for the convolutional encoder used on top of the pre-trained ViT \footnote{we used the features of ViT (token dimensionality 768, 12 heads, patch size 16) pre-trained using DINO, and provided by the timm library \cite{wightman2019pytorch}, with the model name \textit{vit\_base\_patch16\_224.dino}.} features, \cref{slot_dec_vprom} describes the hyperparameters for the decoder, and \cref{slot_attention_pretraining_hyperparams_vprom} describes the pre-training hyperparameters.  After pre-training, we selected the model from the epoch with the lowest mean squared error on the validation set. The slot attention parameters were frozen after pre-training, except for the V-PROM dataset where they were also finetuned on the reasoning task.

We resized the images to $128\times 128$ for ART, SVRT, and CLEVR-ART, and resized the image panels (of which there are 9 per problem) to $80\times80$ for PGM, and $224\times224$ for V-PROM. Pixels were normalized to the range $[0,1]$ for ART, and to $[-1,1]$ for SVRT, CLEVR-ART, and PGM datasets. For V-PROM, we applied a channel-wise mean and standard deviation normalization with values of $[0.485, 0.456, 0.406]$ for the mean, and $[0.229, 0.224, 0.225]$ for the standard deviation corresponding to the RGB channels, after converting the pixel values to the range $[0,1]$. For SVRT, we also applied random horizontal and vertical flips, consistent with prior work \cite{vaishnav2023gamr,webb2023systematic}. 

\begin{table}[t]
\caption{Results on the two tasks (relational-match-to-sample (RMTS) and identity rules (ID)) of the CLEVR-ART dataset. Results reflect test accuracy averaged over 5 trained networks ($\pm$ standard error).}
\label{clevr-art-table}
\vskip 0.15in
\begin{center}
\begin{small}
\begin{sc}
\begin{tabular}{lccr}
\toprule
Model & RMTS & ID \\
\midrule
Slot-CorelNet &49.87$\pm$0.2&24.80$\pm$0.3\\
Slot-ESBN &62.53$\pm$0.1&28.87$\pm$0.7\\
Slot-GAMR &52.56$\pm$0.5&39.83$\pm$0.9\\
Slot-RN &64.79$\pm$0.5&60.27$\pm$0.6\\
Slot-IN &66.72$\pm$3.7&67.22$\pm$1.7\\
Slot-Transformer & 87.54$\pm$0.7 & 78.81$\pm$1.6 \\
 GAMR & 70.40$\pm$5.8 & 74.15$\pm$4.0 \\
 OCRA & 93.34$\pm$1.0 & 77.06$\pm$0.7 \\
 Slot-Abstractor &\textbf{96.34$\pm$0.5}& \textbf{91.61$\pm$0.2}\\
\bottomrule
\end{tabular}
\end{sc}
\end{small}
\end{center}
\vskip -0.1in
\end{table}

\begin{table*}[t]
\caption{Results on different generalization regimes of the PGM dataset. Results reflect test accuracy for 1 trained model for each regime.}
\label{pgm-table}
\vskip 0.15in
\begin{center}
\begin{small}
\begin{sc}
\begin{tabular}{lcccccccccr}
\toprule
Model & Neutral & Intp.  & H.O.A.P. & H.O.T.P. & H.O.Triples & H.O.L-T & H.O.S-C & Extp. \\
\midrule
WReN  $\beta = 0$& 62.6 & 64.4& 27.2& 41.9& 19.0& 14.4& 12.5& 17.2\\
VAE-WReN &64.2& -&36.8 &43.6 &24.6 &- & -&-\\
MXGNet $\beta = 0$ &66.7 & 65.4& 33.6& 43.3 &19.9 & 16.7&16.6 &18.9\\
DCNet &68.6&59.7&-&-&-&-&-&17.8\\
Rel-Base &85.5&-&-&-&-&-&-&22.1\\
MRNet   & 93.4  &68.1  &38.4  &55.3 &25.9 &\textbf{30.1} &\textbf{16.9} &19.2 \\
MLRN &98.0&57.8&-&-&-&-&-&14.9\\
STSN   & \textbf{98.2}  & 78.5 &- &- &- &-&-& 20.4 \\
ARII   & 88.0  &72.0  &50.0  &64.1 &\textbf{32.1} &16.0 &12.7 &29.0 \\
PredRNet &97.4&70.5&\textbf{63.4}&67.8& 23.4 &27.3&13.1 &19.7  \\
Slot-Abstractor &91.5&\textbf{91.6}&\textbf{63.3}&\textbf{78.3}&20.4&16.7&14.3&\textbf{39.3}\\
\bottomrule
\end{tabular}
\end{sc}
\end{small}
\end{center}
\vskip -0.1in
\end{table*}

The hyperparameters for the Abstractor module of the Slot Abstractor are described in \cref{relational_reasoning_hyperparams}. We initialized $\boldsymbol{W_q}$ and $\boldsymbol{W_k}$ (the linear projection matrices used to generate queries and keys) with the same values, but allowed them to vary during training. To generate the output, we took the mean of the $K$ final relational representations $\boldsymbol{s}^{L}$, and passed this through a task-specific output layer. For the same/different ART task and SVRT tasks, this layer had a single unit and a sigmoid activation, and the model was trained with binary cross entropy loss. For the other ART tasks, PGM, and the V-PROM datasets, each candidate answer choice was inserted into the problem before being fed as input to the model, and the output was a linear layer with a single unit. This output was treated as a score for the answer choice in each input. A softmax activation was then applied over the scores for all answer choices, and the model was trained with cross-entropy loss. More experimental details can be found at \cref{more_expt_details}. The code is available at \href{https://github.com/Shanka123/Slot-Abstractor}{https://github.com/Shanka123/Slot-Abstractor}.

\section{Results}
\cref{art-table} shows the results on the four ART tasks. The Slot Abstractor achieved state-of-the-art accuracy on three tasks (same/different, relational-match-to-sample, and distribution-of-3), beating the next best baseline (OCRA) by as much as 9\%. It also demonstrated an average improvement of 3\% over all tasks compared to previous state-of-the-art model (OCRA). 

\cref{svrt-table} shows the results on the two task categories for the SVRT dataset. The Slot Abstractor displayed comparable overall performance with previous state-of-the-art models (GAMR, OCRA), demonstrating a marginal improvement on tasks defined by same/different relations, whereas GAMR performed slightly better on tasks defined by spatial relations. 

\cref{clevr-art-table} shows the results on the two CLEVR-ART tasks. The Slot Abstractor achieved state-of-the-art accuracy beating previous state-of-the-art model (OCRA) by as much as 14\%. This demonstrates that the Slot Abstractor's capacity for systematic generalization can also be extended to abstract reasoning problems with more complicated and realistic visual inputs. 

\cref{pgm-table} shows the results on the PGM generalization regimes. The Slot Abstractor demonstrated notable improvements on many of the OOD generalization regimes (Intp., H.O.T.P., Extp.)  with as much as 21\% compared to the previous state-of-the-art model (PredRNet), and an average improvement of around 5\% over all the regimes. It is also worth noting that OCRA, previously the state-of-the-art model for ART, SVRT, and CLEVR-ART datasets, could not be trained on the PGM dataset due to memory issues. It was not possible to train the model on an A100 GPU with 80GB memory, even when using a batch size of 1 (given $144$ slots, OCRA would need to process $144^2\approx20$k relational embeddings). This highlights the benefits of the Slot Abstractor's improved complexity ($O(N^2)$ vs. $O(N^4)$ in OCRA), allowing it to be effectively applied to problems involving a much larger number of objects.

On V-PROM, the Slot Abstractor displayed state-of-the-art performance achieving 67.7\% test accuracy, compared to 51.2\% by the previous best model, Relation Net. This demonstrates that the Slot Abstractor's systematic generalization capacity can also be extended to abstract visual reasoning problems involving real-world images.

\begin{table}[t]
\caption{Results on the PGM Neutral regime when trained on a subset (20\% vs. 100\%) of the training data. Results reflect test accuracy for 1 trained model.}
\label{sample-efficiency-table}
\vskip 0.15in
\begin{center}
\begin{small}
\begin{sc}
\begin{tabular}{lcccr}
\toprule
Model & \multicolumn{2}{c}{\% training data} \\
 & 20\% & 100\% \\
\midrule
 STSN & 53.35 & \textbf{98.2}   \\
 Slot-Abstractor & \textbf{71.25}& 91.5  \\
 
\bottomrule
\end{tabular}
\end{sc}
\end{small}
\end{center}
\vskip -0.1in
\end{table}

To evaluate the Slot Abstractor's sample efficiency, we trained it on 20\% of the training data in the PGM neutral regime. We also compared it to the state-of-the-art model in the neutral regime, STSN  \cite{mondal2023stsn}, using the publicly available code \footnote{\href{https://github.com/Shanka123/STSN}{https://github.com/Shanka123/STSN}}. As done in the original implementation, slot attention parameters were initialized using the same pre-trained slot attention as used in the Slot Abstractor, but then fine-tuned while training on the downstream task. \cref{sample-efficiency-table} shows the results. When trained on only 20\% of the training data, the Slot Abstractor significantly outperforms STSN, demonstrating the Slot Abstractor's superior sample efficiency.  

\begin{table*}[t]
\caption{Ablation study on tasks of the ART dataset. Results reflect test accuracy averaged over 10 trained networks($\pm$ standard error). }
\label{art-ablation-table}
\vskip 0.15in
\begin{center}
\begin{small}
\begin{sc}
\begin{tabular}{lccccr}
\toprule
Model & Same/Different & Relational-match-to-sample  & Distribution-of-3 & Identity rules \\
\midrule
Slot-Abstractor &\textbf{96.36$\pm$ 0.4}& \textbf{91.64$\pm$ 1.6}&\textbf{95.22$\pm$ 0.4}&\textbf{96.41$\pm$0.4}  \\
w/o Slot Attention &89.29$\pm$0.8&49.85$\pm$0.2&89.10$\pm$1.2&86.09$\pm$1.3\\
w/o Self Attention &\textbf{95.26$\pm$0.9}&85.41$\pm$2.0&93.83$\pm$0.6&92.58$\pm$0.7\\
w/o Factorized Reps. &77.24$\pm$2.8&50.02$\pm$0.1&64.4$\pm$1.5&85.08$\pm$1.5\\
Replace RCA with CA &80.04$\pm$1.4&58.24$\pm$1.5&43.90$\pm$1.3&60.02$\pm$1.0\\
\bottomrule
\end{tabular}
\end{sc}
\end{small}
\end{center}
\vskip -0.1in
\end{table*}

\subsection{Ablation Study}
\label{ablation_study}
To understand the importance of each of the Slot Abstractor's major components, we performed an ablation study using the ART dataset (\cref{art-ablation-table}). First, we removed Slot Attention, instead dividing the feature map $\boldsymbol{feat}$ into a $4\times4$ grid (treating each location within the feature map as an `object'), and trained the model end to end. We noticed a significant drop in performance for the relational-match-to-sample and same/different tasks. This demonstrates the importance of using object-centric representations (even when pre-trained DINO ViT features are used, as described in \ref{dinosaur_slot_attention_ablation}). Second, we removed different components from the Abstractor module. In one case, we removed the self-attention and the subsequent feedforward layer. This impaired performance on the relational-match-to-sample, distribution-of-3, and identity rules tasks, where the ablation of self-attention resulted in 4-7\% drop in test accuracy. This demonstrates the importance of modeling higher-order relations through self-attention. In another case, we ablated the factorized slot representations, using the standard slot representations instead of separate feature and position embeddings. The relation embeddings $\boldsymbol{s}^{l=0}$ were initialized using learned parameters. We observed a significant drop in performance (as much as 40\%) for all tasks except identity rules. This demonstrates the importance of using factorized feature and position embeddings, which allow the Slot Abstractor to keep track of the correspondence between relations and objects. Finally, we replaced the relational cross-attention ($\operatorname{RCA}$) with standard cross-attention ($\operatorname{CA}$), where keys and values are formed from feature embeddings, and queries are formed from position embeddings, thereby removing the relational bottleneck. This significantly impaired performance for all tasks (by as much as 52\%). This reflects the centrally important role of the relational bottleneck as an inductive bias underlying the Slot Abstractor's capacity for generalization.

The removal of slot attention (Row 2 of \cref{art-ablation-table}) and the replacement of relational cross-attention with standard cross-attention (Row 5 of \cref{art-ablation-table}), resulted in a significant drop in performance, which demonstrates the importance of combining slot attention with relational cross-attention in Abstractors. The removal of self-attention and the subsequent feedforward layer after relational cross-attention (Row 3 of \cref{art-ablation-table}) and the use of standard slot representations without factorizing into position and feature specific embeddings (Row 4 of \cref{art-ablation-table}) explored different design choices of combining slot attention with the relational cross-attention in Abstractors, both of which significantly underperformed compared to the proposed method, thereby demonstrating that these design choices are essential for obtaining strong performance.

\subsection{Visualization of Abstractor's Relation Embeddings}

\begin{figure}[h!]
\vskip 0.2in
\begin{center}
\centerline{\includegraphics[width=0.85\columnwidth]{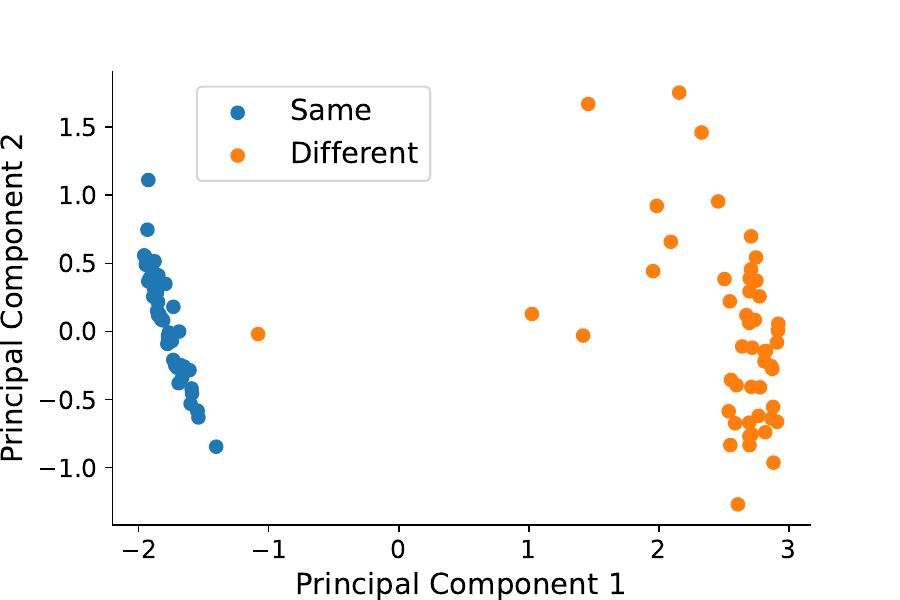}}
\caption{\textbf{Visualization of Abstractor's Relation Embeddings.} The output of the first head of relational cross attention, after projecting to the first two principal components for 100 examples from the test set of same/different ART task. Two different clusters are formed corresponding to problems with the same and different relation among the objects.}
\label{rca_head_relations_plot}
\end{center}
\vskip -0.2in
\end{figure}

To demonstrate that the Abstractor can capture the different relations present in the tasks, we performed a qualitative analysis of the relational cross-attention head. Specifically, we visualized the output of the relational cross-attention corresponding to the first head ($\boldsymbol{\tilde{s}}^{1}$ in \cref{symbol_eq}) for 100 examples from the test set of same/different ART task, with equal number of examples belonging to same and different category. We applied principal component analysis (PCA) to the output features after taking the mean over the $K$ slots, to reduce it to 2 dimensions. \cref{rca_head_relations_plot} shows that the two-dimensional output form two different clusters corresponding to images belonging to the same and different categories, indicating that the first relational cross-attention head was able to capture the same/different relation.

\section{Conclusion and Future Directions}
 In this work, we developed Slot Abstractors, by combining slot-based object-centric encoding mechanisms with Abstractors, a recently proposed approach for implementing the relational bottleneck using a relational variant of cross-attention. We evaluated the Slot Abstractor on five abstract visual reasoning tasks including one involving real-world images, finding that it achieves overall state-of-the-art systematic generalization of learned abstract rules, and can also be scaled to problems containing a large number of objects and multiple rules.

Finally, we discuss some limitations of our work and potentially interesting directions for future work. First, although Slot Abstractors demonstrated state-of-the-art performance in many of the OOD generalization regimes of PGM (and achieved state-of-the-art average performance across all regimes), there is still significant room for improvement in some of these regimes. Even for some regimes on which the Slot Abstractor achieved state-of-the-art performance, that performance was sometimes well below what would be expected of human reasoners (e.g., especially for the `Extrapolation' regime). Second, the number of slots in slot-based models is fixed, which may pose a challenge for settings in which there are large variations in the number of objects that must be processed.
Recent work \cite{lowe2023rotating,stanic2023contrastive} has built methods addressing this limitation by doing away with slots entirely, and future work might explore how such non-slot-based methods can be integrated with relational inductive biases. Finally, future work could also improve the quadratic complexity of Slot Abstractors by using more efficient attention mechanisms \cite{dao2022flashattention,dao2023flashattention}.

% Acknowledgements should only appear in the accepted version.
\section*{Acknowledgements}
Shanka Subhra Mondal was supported by Office of Naval Research grant N00014-22-1-2002 during
the duration of this work. We would like to thank the reviewers for their valuable feedback, and the Princeton Research Computing, especially William
G. Wischer and Josko Plazonic, for their help with scheduling training jobs on the Princeton University
Della cluster.

\section*{Impact Statement}

This paper presents work the goal of which is to advance the field of Machine Learning. There are many potential societal consequences of our work, none of which we feel must be specifically highlighted here.

% In the unusual situation where you want a paper to appear in the
% references without citing it in the main text, use \nocite
\nocite{langley00}

\bibliography{example_paper}
\bibliographystyle{icml2024}

%%%%%%%%%%%%%%%%%%%%%%%%%%%%%%%%%%%%%%%%%%%%%%%%%%%%%%%%%%%%%%%%%%%%%%%%%%%%%%%
%%%%%%%%%%%%%%%%%%%%%%%%%%%%%%%%%%%%%%%%%%%%%%%%%%%%%%%%%%%%%%%%%%%%%%%%%%%%%%%
% APPENDIX
%%%%%%%%%%%%%%%%%%%%%%%%%%%%%%%%%%%%%%%%%%%%%%%%%%%%%%%%%%%%%%%%%%%%%%%%%%%%%%%
%%%%%%%%%%%%%%%%%%%%%%%%%%%%%%%%%%%%%%%%%%%%%%%%%%%%%%%%%%%%%%%%%%%%%%%%%%%%%%%
\newpage
\appendix
\onecolumn
\section{Appendix}

\subsection{Algorithm for Slot Abstractors}
\begin{algorithm}[h!]
   \caption{Slot Abstractors: The inputs are the feature map obtained by passing an image $\boldsymbol{x}$ through a convolutional encoder, a position code (encoding the four cardinal directions) which is passed through a linear layer to generate positional embeddings, and $K$ slots which are initialized from shared mean and variance parameters.}
   \label{alg:slot abstractors alg}
\begin{algorithmic}
   \STATE {\bfseries Inputs:} feature map $\boldsymbol{feat} \in \mathbb{R}^{N \times D}$ ;  positional embeddings $\boldsymbol{pos} \in \mathbb{R}^{N \times D}$ ; $\boldsymbol{slots} \sim Normal (\mu,\sigma) \in \mathbb{R}^{K \times D}, \mu \in   \mathbb{R}^D, \sigma \in \mathbb{R}^D$
    \STATE {\bfseries Parameters:} $q, k, v$ linear projection matrices for slot attention, relational cross-attention, self-attention; GRU; FeedForward; slot attention iterations $T$; attention heads $H$; Abstractor layers $L$
    \STATE $\boldsymbol{inputs} = $ FeedForward($\boldsymbol{feat} +\boldsymbol{pos}$)
    \FOR{$t=1$ {\bfseries to} $T$}
 
\STATE $\boldsymbol{attn}_t = \operatorname{softmax} (\frac{1}{\sqrt{D}} \operatorname{k}(\boldsymbol{inputs}) \cdot \operatorname{q}(\boldsymbol{slots})^{\top}, axis =$ `$slots$'$)$

   \STATE $\boldsymbol{updates} = \boldsymbol{attn}_t \cdot \operatorname{v}(\boldsymbol{inputs})$ 
   \STATE $\boldsymbol{slots}$ =  GRU(inputs = $\boldsymbol{updates}$ , hidden state = $\boldsymbol{slots}$ )
   \STATE $\boldsymbol{slots} =$ FeedForward$( \boldsymbol{slots}) + \boldsymbol{slots}$
    \ENDFOR
    
     \STATE $\boldsymbol{z}_{k} = \boldsymbol{attn}_{T} \operatorname{flatten}(\boldsymbol{feat})[k]$

     \STATE $\boldsymbol{m}_{k} = \boldsymbol{attn}_{T} \operatorname{flatten}(\boldsymbol{pos})[k]$
     \STATE $\boldsymbol{z} = \{\boldsymbol{z}_{1},... \boldsymbol{z}_{K}\}$
     \STATE $\boldsymbol{m} = \{\boldsymbol{m}_{1},... \boldsymbol{m}_{K}\}$
   \STATE Initialize $\boldsymbol{s}^{0}$ = $\boldsymbol{m}$
   \FOR{$l=1$ {\bfseries to} $L$}
   \STATE $\boldsymbol{s}^{l}$ = RelationalCrossAttention($\boldsymbol{z}$, $\boldsymbol{s}^{l-1}$) + $\boldsymbol{s}^{l-1}$
   
   \STATE $\boldsymbol{s}^{l}$ = FeedForward($\boldsymbol{s}^{l}$) + $\boldsymbol{s}^{l}$

   \STATE $\boldsymbol{s}^{l}$ = SelfAttention($\boldsymbol{s}^{l}$) + $\boldsymbol{s}^{l}$

   \STATE $\boldsymbol{s}^{l}$ = FeedForward($\boldsymbol{s}^{l}$) + $\boldsymbol{s}^{l}$
   \ENDFOR

   \STATE {\bfseries Output:} $\boldsymbol{s}^{ L}$
   
\end{algorithmic}
\end{algorithm}

\subsection{More Experimental Details}
\label{more_expt_details}
For ART and CLEVR-ART tasks, we used a learning rate of $8e-5$, batch size of 16, and the number of training epochs are described in \cref{art_clevrart_epochs}. For SVRT, we used a learning rate of $4e-5$, a batch size of 32, and trained for 2000 epochs on each task. We used a single A100 GPU with 80GB memory for training on ART, SVRT, and CLEVR-ART tasks. For training on PGM regimes, we used 4 A100 GPUs with 80GB memory each, a learning rate of $8e-5$, and a batch size of 24 per GPU. The number of training epochs for each PGM generalization regime is displayed in \cref{pgm_epochs}. For training on the V-PROM neutral regime, we also used 4 A100 GPUs with 80GB memory each, a learning rate of $8e-5$, and a batch size of 12 per GPU, and trained for 30 epochs. For optimization, we used the ADAM \cite{kingma2014adam} optimizer, and for implementation, we used the PyTorch library \cite{paszke2017automatic}. 

\subsection{Importance of Slot Attention in DINOSAUR}
\label{dinosaur_slot_attention_ablation}
To justify the importance of the slot attention component, when the DINOSAUR model is used with the Abstractor, we conducted an ablation experiment on the V-PROM neutral regime, by removing the slot attention component from the DINOSAUR model, and used the pre-trained DINO ViT features with the Abstractor. Since, it wasn’t possible to train the ablation model using the patchwise DINO ViT features on an A100 GPU of 80GB memory even with a batch size of 1 (given 9 image panels and 196 patches for each panel, the size of the matrix which models relations between the features would be $1764\times1764$), and hence we used the CLS output of the DINO ViT features, as feature embeddings $\boldsymbol{z}$ to Abstractor. On V-PROM, the ablation model achieved a test accuracy of 30.01\%, compared to 67.7\% by our proposed method. This result suggests the importance of slot attention as an effective way of finding a low-dimensional representation, with slots bound to objects in the image.

\subsection{Datasets}

\begin{figure}[h!]
\centering
\includegraphics[width=0.9\textwidth]{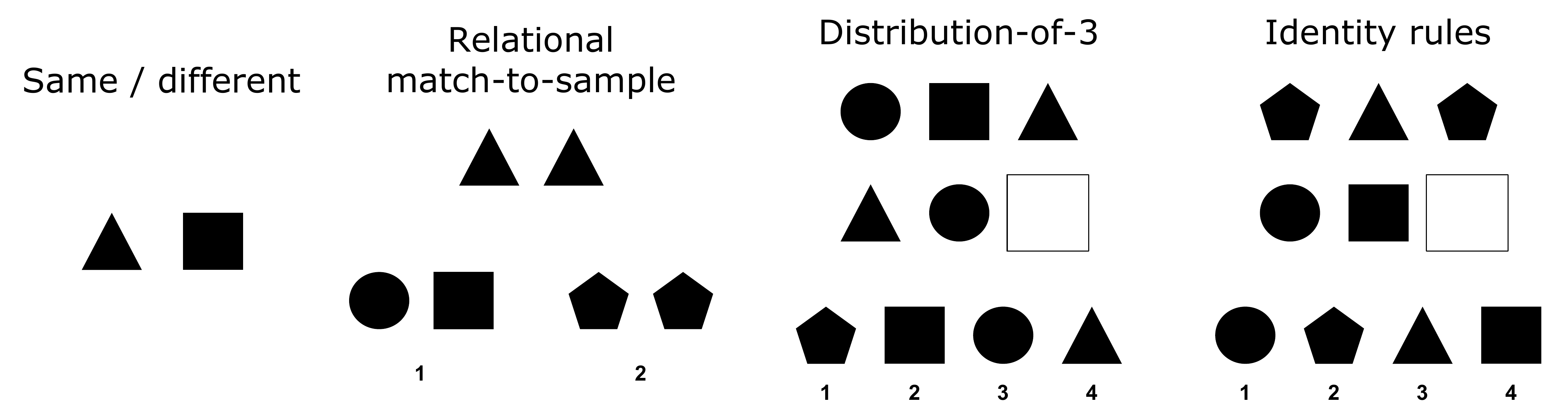} 
\caption{\textbf{Abstract Reasoning Tasks (ART) Dataset}. The `same/different' task, requires identifying whether two objects are the same or different. The `relational-match-to-sample' task requires selecting a pair of objects, out of two pairs called the target objects that has the same relation, as the relation (`same' or `different') among a source pair of objects. We presented the problem as a $2\times2$ array format, with the source pair of objects in the top row, and a target pair in the bottom row (separate images for each target pair). 
 In the `distribution-of-3' task, the first row contains a set of three objects, and the second row contains an incomplete set. The task is to select the missing object from a set of four choices. We presented the problem as a $2\times3$ array format, with one of the choices inserted in the bottom right cell (separate images for each choice). In the `identity rules' task, the first row contains three objects that follow an abstract pattern (ABA, ABB, or AAA), and the task is to select the choice that would result in the same relation being instantiated in the second row. We presented the problems in the same format as the `distribution of-3' task.}  
\label{art_dataset}
\end{figure}

\begin{table}[h!]
\caption{Number of training examples for ART tasks.}
\label{art_examples}
\vskip 0.15in
\begin{center}
\begin{small}
\begin{sc}
\begin{tabular}{cccc}
\toprule
 Same/Different & Relational-match-to-sample  & Distribution-of-3 & Identity rules  \\
\midrule
40& 480 & 360 &8640  \\

\bottomrule
\end{tabular}
\end{sc}
\end{small}
\end{center}
\vskip -0.1in
\end{table}

% \pagebreak

\begin{figure}[h!]
    \centering
    \subfigure[Examples of same/different task.]{\includegraphics[width=0.4\linewidth]{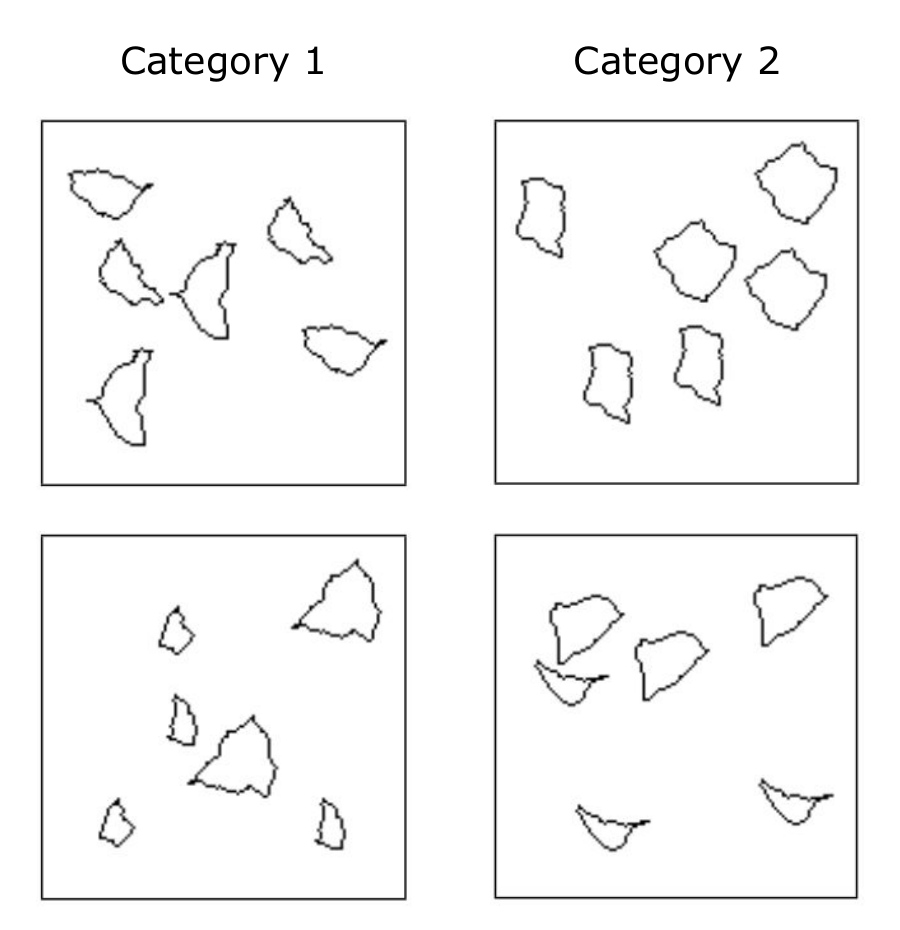}\label{svrt_sd_example}}%
    \subfigure[Examples of spatial relation task.]{\includegraphics[width=0.4\linewidth]{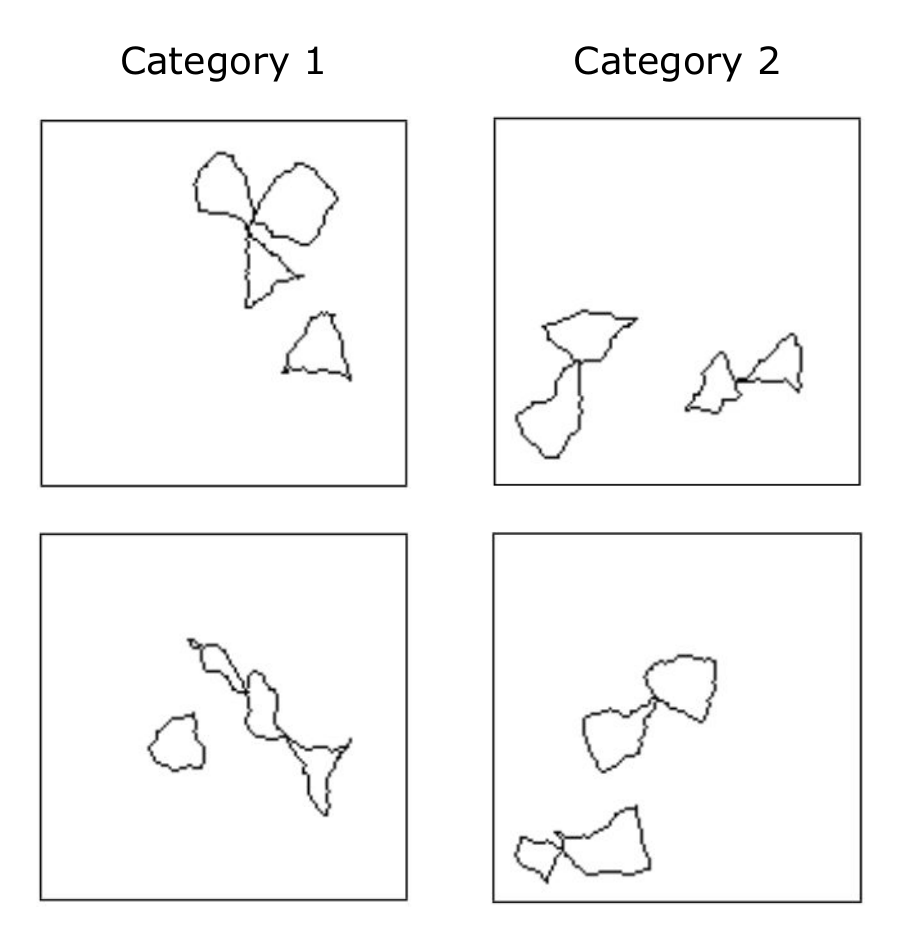}\label{svrt_sr_example}}%
    \caption{\textbf{Synthetic Visual Reasoning Test (SVRT) Dataset.} \textbf{(a)} Examples of task depicting same/different relation. Each row shows an example from each of the two categories. In category 1 there are three sets of two identical shapes. In category 2 there are two sets of three identical shapes. \textbf{(b)} Examples of task depicting spatial relation. Each row shows an example from each of the two categories. In category 1 three out of four shapes are touching each other. In category 2 there are two sets of two shapes touching each other. }%
    \label{svrt_dataset}
\end{figure}

\begin{figure}[h!]
\centering
\includegraphics[width=0.9\textwidth]{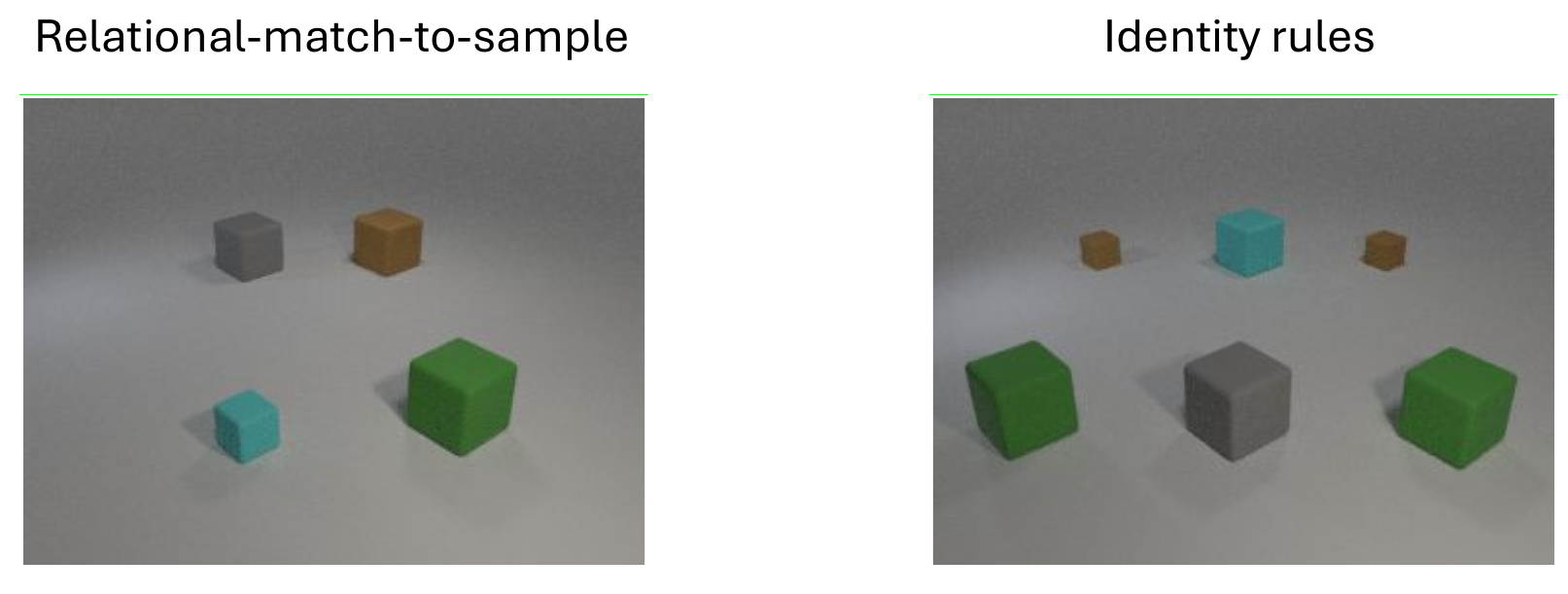} 
\caption{\textbf{CLEVR-ART Dataset}. \textbf{Relational-match-to-sample:} An example depicting `different' relation between the source pair (back row) of objects and the target pair (front row) of objects. Problems were presented in the same format as the `relational match-to-sample' task of the ART dataset. \textbf{Identity rules:} An example problem depicting ABA rule among the back row and front row of objects. Problems were presented in the same format as the `identity rules' task of the ART dataset. }  
\label{clevr_art_dataset}
\end{figure}

\begin{figure}[h!]
\centering
\includegraphics[width=0.9\textwidth]{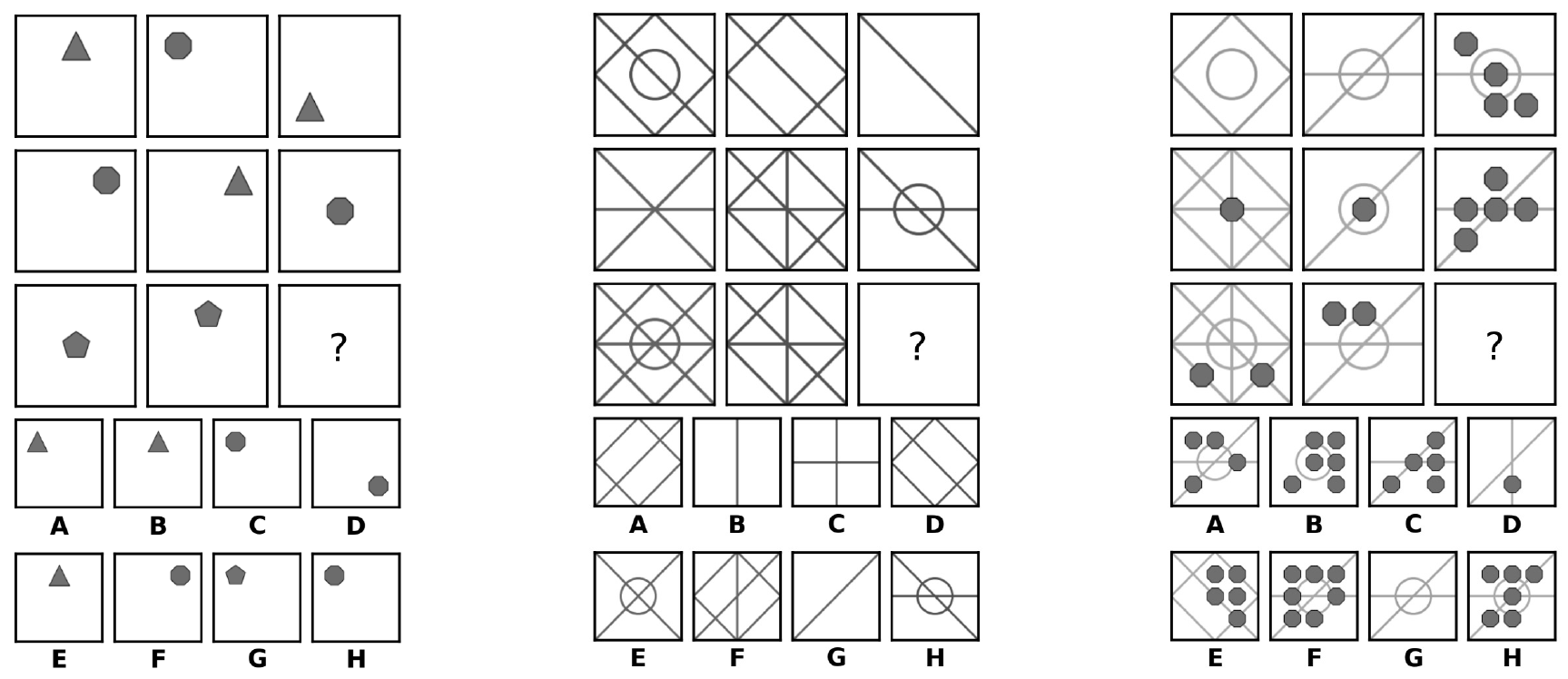} 
\caption{\textbf{Procedurally Generated Matrices (PGM) Dataset}. Examples of problems with only shapes (left), only lines (middle), and both shapes and lines (right). In each problem, the task is to identify the abstract pattern among the image panels in the first two rows and/or columns of the matrix, and use that pattern to fill the panel in the third row and column from a set of eight choices.}  
\label{pgm_dataset}
\end{figure}
\pagebreak

\subsubsection{PGM OOD Regimes}
\label{pgm_ood_regimes}
Each matrix problem in PGM is defined by the abstract structure $\mathcal{S} = \{[r, o ,a]: r \in \mathcal{R}, o \in \mathcal{O}, a \in \mathcal{A} \}$, where $\mathcal{R} = $ \{progression, XOR, AND, OR, consistent union\} are the set of rules, $\mathcal{O}=$ \{shape, line\} are the set of objects, and $\mathcal{A}=$ \{size, type, position, color, number\} are the set of attributes. Problems consist of multiple rules (upto 4) governed by triples $[r, o ,a]$. The PGM dataset contains seven OOD regimes of varying difficulty. The interpolation regime involves training on even-indexed feature values for color and size attributes, and testing on all odd-indexed values, whereas the extrapolation regime involves training on the lower half of feature values and testing on the upper half of feature values. In the heldout attribute shape color (H.O.S-C) regime, the training set problems doesn't contain shape objects and color attribute, whereas all problems in the test set contain at least shape objects and color attribute. Similarly in the heldout attribute line type (H.O.L-T) regime, the training set problems doesn't contain line objects and type attribute. The heldout triples (H.O.Triples) regime, involves randomly holding out seven triples out of 29 possible unique triples $[r, o, a]$ for test set problems. In heldout triple pairs (H.O.T.P.) regime, the problems in both the training and test set contain at least two triples, and out of 400 possible pairs of triples, randomly chosen 40 pairs of triples are used only for test set problems. The heldout arribute pairs (H.O.A.P.) regime, involves holding out 4 attribute pairs out of possible 20 pairs for test set problems.

\begin{figure}[h!]
\centering
\includegraphics[width=0.9\textwidth]{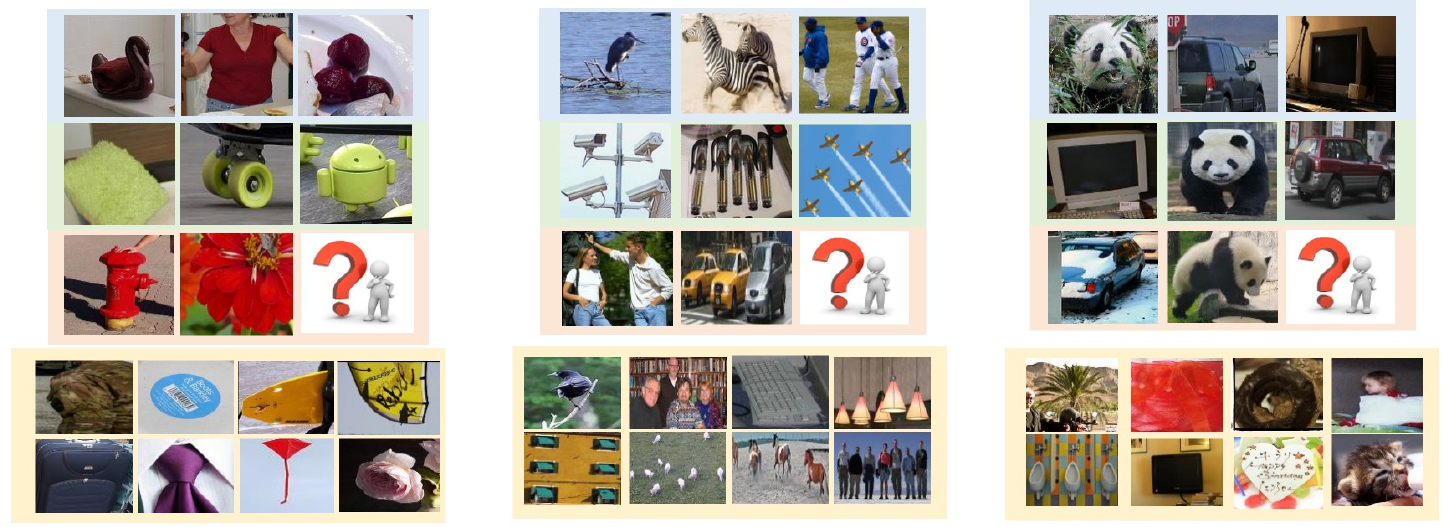} 
\caption{\textbf{Visual Progressive Matrices (V-PROM) Dataset}. Example problems of Raven's Progressive Matrices involving real-world images. Similar to PGM dataset, the task is to identify the abstract pattern among the images in the first two rows and/or columns of the $3\times3$ matrix, and use that pattern to fill the missing image in the third row, by selecting one among a set of eight choices.} 
\label{vprom_dataset}
\end{figure}

\subsection{Visualization of Slot-wise Reconstruction}

\begin{figure}[h!]
\centering
\includegraphics[width=1\textwidth]{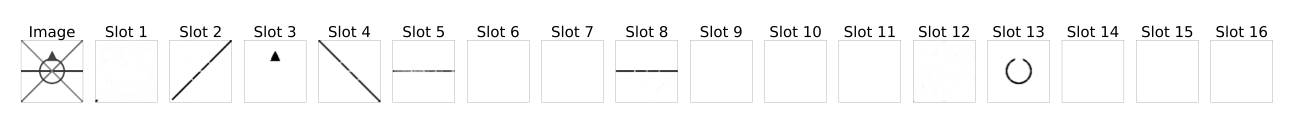} 
\caption{Visualization of the individual slot-wise reconstruction for all the 16 slots for an image panel from the PGM dataset.}  
\label{pgm_slot_recon_ex1}
\end{figure}

\begin{figure}[h!]
\centering
\includegraphics[width=1\textwidth]{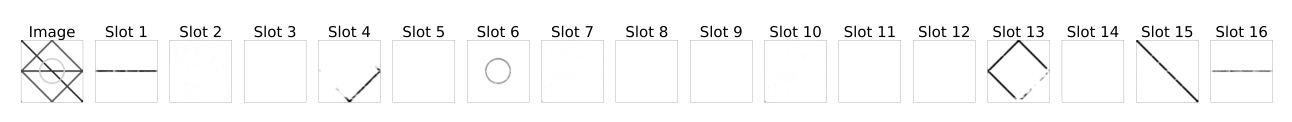} 
\caption{Visualization of the individual slot-wise reconstruction for all the 16 slots for an image panel from the PGM dataset.}  
\label{pgm_slot_recon_ex2}
\end{figure}
\subsection{Training Details and Hyperparameters}

%%%%%%%%%%%%%%%%%%%%%%%%%%%%%%%%%%%%%%%%%%%%%%%%%%%%%%%%%%%%%%%%%%%%%%%%%%%%%%%
%%%%%%%%%%%%%%%%%%%%%%%%%%%%%%%%%%%%%%%%%%%%%%%%%%%%%%%%%%%%%%%%%%%%%%%%%%%%%%%
\begin{table}[h!]
\caption{CNN encoder hyperparameters for PGM dataset.}
\label{cnn_enc_pgm}
\vskip 0.15in
\begin{center}
\begin{small}
\begin{sc}
\begin{tabular}{lcccccr}
\toprule
Type & Channels & Activation  & Kernel size & Stride & Padding \\
\midrule
2D Conv   & 32& ReLU & 5$\times$5 & 1& 2\\
2D Conv   & 32& ReLU & 5$\times$5 & 1& 2\\
2D Conv   & 32& ReLU & 5$\times$5 & 1& 2\\
2D Conv   & 32& ReLU & 5$\times$5 & 1& 2\\
Position Embedding &- & -&- &-&-& \\
Flatten &- & -&- &-&-& \\
Layer Norm &- & -&- &-&-& \\
1D Conv   & 32& ReLU & 1 & 1& 0\\
1D Conv   & 32& - & 1 & 1& 0\\
\bottomrule
\end{tabular}
\end{sc}
\end{small}
\end{center}
\vskip -0.1in
\end{table}

\begin{table}[h!]
\caption{Slot decoder hyperparameters for PGM dataset.}
\label{slot_dec_pgm}
\vskip 0.15in
\begin{center}
\begin{small}
\begin{sc}
\begin{tabular}{lcccccr}
\toprule
Type & Channels & Activation  & Kernel size & Stride & Padding \\
\midrule
Spatial Broadcast &- & -&- &-&-& \\
Position Embedding &- & -&- &-&-& \\
2D Conv   & 32& ReLU & 5$\times$5 & 1& 2\\
2D Conv   & 32& ReLU & 5$\times$5 & 1& 2\\
2D Conv   & 32& ReLU & 5$\times$5 & 1& 2\\
2D Conv   & 2& - & 3$\times$3 & 1& 1\\
\bottomrule
\end{tabular}
\end{sc}
\end{small}
\end{center}
\vskip -0.1in
\end{table}

\begin{table}[h!]
\caption{CNN encoder hyperparameters for V-PROM dataset.}
\label{cnn_enc_vprom}
\vskip 0.15in
\begin{center}
\begin{small}
\begin{sc}
\begin{tabular}{lcccccr}
\toprule
Type & Channels & Activation  & Kernel size & Stride & Padding \\
\midrule
Layer Norm &- & -&- &-&-& \\
1D Conv   & 768& ReLU & 1 & 1& 0\\
1D Conv   & 256& - & 1 & 1& 0\\
\bottomrule
\end{tabular}
\end{sc}
\end{small}
\end{center}
\vskip -0.1in
\end{table}

\begin{table}[h!]
\caption{Slot decoder hyperparameters for V-PROM dataset.}
\label{slot_dec_vprom}
\vskip 0.15in
\begin{center}
\begin{small}
\begin{sc}
\begin{tabular}{lcccccr}
\toprule
Type & Channels & Activation  & Kernel size & Stride & Padding \\
\midrule
Spatial Broadcast &- & -&- &-&-& \\
Position Embedding &- & -&- &-&-& \\
1D Conv   & 2048& ReLU & 1 & 1& 0\\
1D Conv   & 2048& ReLU & 1 & 1& 0\\
1D Conv   & 2048& ReLU & 1 & 1& 0\\
1D Conv   & 769& - & 1 & 1& 0\\
\bottomrule
\end{tabular}
\end{sc}
\end{small}
\end{center}
\vskip -0.1in
\end{table}

\begin{table}[h!]
\caption{Slot attention pretraining hyperparameters for PGM dataset.}
\label{slot_attention_pretraining_hyperparams}
\vskip 0.15in
\begin{center}
\begin{small}
\begin{sc}
\begin{tabular}{lcr}
\toprule
 & \\
\midrule
Slot dimensionality $D$ & 32\\
No of iterations $T$ & 3\\
No of slots $K$ & 16\\
Batch size & 16\\
Learning rate & $4e-4$\\
Learning rate warmup steps & 75000 \\
Epochs before learning rate decay & 5\\ 
Learning rate decay rate & 0.5\\ 
Learning rate decay steps & 100000 
\\
Epochs & 8\\

\bottomrule
\end{tabular}
\end{sc}
\end{small}
\end{center}
\vskip -0.1in
\end{table}

\begin{table}[h!]
\caption{Slot attention pretraining hyperparameters for V-PROM dataset.}
\label{slot_attention_pretraining_hyperparams_vprom}
\vskip 0.15in
\begin{center}
\begin{small}
\begin{sc}
\begin{tabular}{lcr}
\toprule
 & \\
\midrule
Slot dimensionality $D$ & 256\\
No of iterations $T$ & 3\\
No of slots $K$ & 11\\
Batch size & 48\\
Learning rate & $4e-4$\\
Learning rate warmup steps & 10000 \\
Epochs & 80\\

\bottomrule
\end{tabular}
\end{sc}
\end{small}
\end{center}
\vskip -0.1in
\end{table}

\begin{table}[h!]
\caption{Hyperparameters of the Abstractor module of the Slot Abstractor. For both multi-head relational cross-attention and multi-head self-attention, $H$ is the number of heads, $L$ is the number of layers, $D_{head}$ is the dimensionality of each head. $D_{MLP}$ is the dimensionality of the MLP hidden layer in the feedforward network.}
\label{relational_reasoning_hyperparams}
\vskip 0.15in
\begin{center}
\begin{small}
\begin{sc}
\begin{tabular}{lccccr}
\toprule
 & ART & SVRT  & CLEVR-ART & PGM & V-PROM \\
\midrule
$H$& 8&8 &8 &8 & 8\\
$L$& 6& 24& 24& 24 & 24\\
$D_{head}$&64 &64 &64 & 32 & 256 \\
$D_{MLP}$&512 &512 & 512& 512 &  512\\
Dropout &0&0&0&0 &0.1\\
\bottomrule
\end{tabular}
\end{sc}
\end{small}
\end{center}
\vskip -0.1in
\end{table}

\begin{table}[h!]
\caption{Number of training epochs for ART and CLEVR-ART tasks.}
\label{art_clevrart_epochs}
\vskip 0.15in
\begin{center}
\begin{small}
\begin{sc}
\begin{tabular}{lccccccccr}
\toprule
 & Same/Different & Relational-match-to-sample  & Distribution-of-3 & Identity rules  \\
\midrule
ART & 600& 400 & 400 &100  \\
CLEVR-ART &- & 50& -& 200  \\
\bottomrule
\end{tabular}
\end{sc}
\end{small}
\end{center}
\vskip -0.1in
\end{table}

\begin{table}[h!]
\caption{Number of training epochs for PGM generalization regimes.}
\label{pgm_epochs}
\vskip 0.15in
\begin{center}
\begin{small}
\begin{sc}
\begin{tabular}{lccccccccr}
\toprule
 & Neutral & Interpolation  & H.O.A.P. & H.O.T.P. & H.O.Triples & H.O.L-T & H.O.S-C & Extrapolation \\
\midrule
PGM &73 &50 &90 & 90& 90 & 65& 65& 115 \\

\bottomrule
\end{tabular}
\end{sc}
\end{small}
\end{center}
\vskip -0.1in
\end{table}

\pagebreak

%%%%%%%%%%%%%%%%%%%%%%%%%%%%%%%%%%%%%%%%%%%%%%%%%%%%%%%%%%%%%%%%%%%%%%%%%%%%%%%
%%%%%%%%%%%%%%%%%%%%%%%%%%%%%%%%%%%%%%%%%%%%%%%%%%%%%%%%%%%%%%%%%%%%%%%%%%%%%%%

\end{document}